\pdfoutput=1

\documentclass[11pt]{article}

\usepackage[final]{acl}

\usepackage{times}
\usepackage{latexsym}
\usepackage{graphicx}
\usepackage{multirow}
\usepackage{multicol}
\usepackage{amsmath}
\usepackage{subfig}
\usepackage{booktabs}

\usepackage{tabularx}
\usepackage{tabu}

\usepackage[T1]{fontenc}

\usepackage[utf8]{inputenc}

\usepackage{microtype}

\title{Investigating Data Contamination in Modern Benchmarks \\for Large Language Models}
\author{Chunyuan Deng$^1$ \quad Yilun Zhao$^2$ \quad Xiangru Tang$^2$ 
\bf{\quad Mark Gerstein$^2$ \quad Arman Cohan$^{2,3}$} \vspace{4pt}\\
$^1$Georgia Institute of Technology \quad $^2$Yale University \quad $^3$Allen Institute for AI\\
\texttt{cdeng73@gatech.edu} \quad \texttt{arman.cohan@yale.edu}
}

\begin{document}
\maketitle
\begin{abstract}
Recent observations have underscored a disparity between the inflated benchmark scores and the actual performance of LLMs, raising concerns about potential contamination of evaluation benchmarks. This issue is especially critical for closed-source models and certain open-source models where training data transparency is lacking. In this paper we study data contamination by proposing two methods tailored for both open-source and proprietary LLMs. We first introduce a retrieval-based system to explore potential overlaps between evaluation benchmarks and pretraining corpora. We further present a novel investigation protocol named \textbf{T}estset \textbf{S}lot Guessing (\textit{TS-Guessing}), applicable to both open and proprietary models. This approach entails masking a wrong answer in a multiple-choice question and prompting the model to fill in the gap. Additionally, it involves obscuring an unlikely word in an evaluation example and asking the model to produce it.
We find that certain commercial LLMs could surprisingly guess the missing option in various test sets. Specifically, in the MMLU benchmark, ChatGPT and GPT-4 demonstrated an exact match rate of 52\% and 57\%, respectively, in guessing the missing options in benchmark test data. We hope these results underscore the need for more robust evaluation methodologies and benchmarks in the field.
\end{abstract}

\section{Introduction}
\begin{figure*}[!ht]
\begin{center}
        \centering 
        \includegraphics[width=\linewidth]{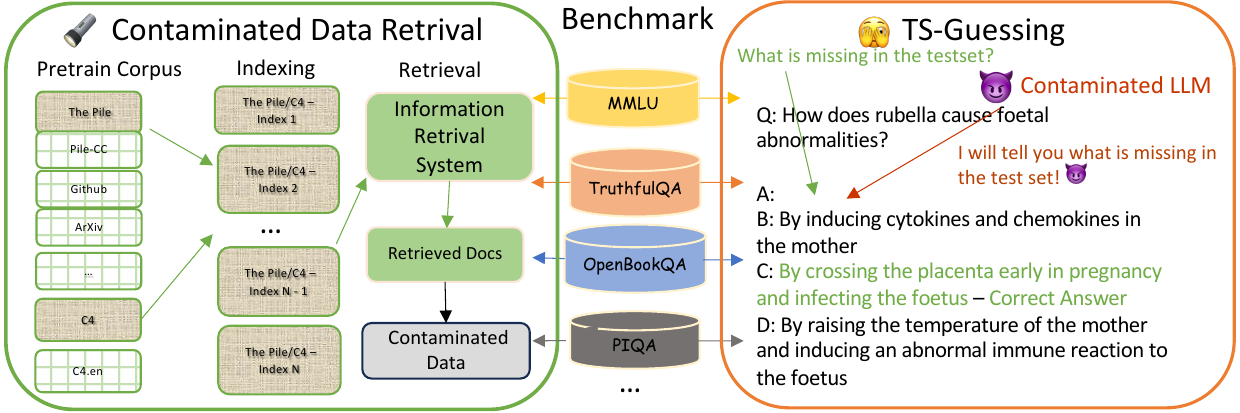}
        \caption{Illustration of our method for identifying data contamination in modern benchmarks. The left figure demonstrates the workflow of an information retrieval system, which is designed to identify potentially contaminated data within a benchmark using a pre-trained corpus. On the right is \textbf{TS-Guessing}, a new investigative approach for potential contamination detection. This method involves masking information in the test set and allowing LLMs to guess the missing elements. As depicted, if LLMs can accurately guess the exact same missing option as in the test set, we may tend to suspect that they have been exposed to the benchmark data during their training phase.}
        \label{fig:flowchart}

\end{center}
\end{figure*}
Large language models (LLMs) have demonstrated exceptional performance across a wide range of NLP tasks, and the NLP community has witnessed the emergence of several impressive LLMs. Notably, there are robust proprietary LLMs, including the GPT-*~\cite{brown2020language, openai2023gpt4}, Claude~\cite{claude}, and Bard~\cite{bard}, among others. In addition to these proprietary models, there are numerous open-source LLMs, such as Llama~\cite{touvron2023llama, touvron2023llama2}, MPT~\cite{lin2023mpt}, Falcon~\cite{mei2022falcon}, and Mistral~\cite{jiang2023mistral}.
However, with the increasing compute scale (including data) used to train these models, concerns have arisen regarding the extensive use of crawled web data, often at a terabyte scale. This extensive training data may, in turn, potentially include instances of evaluation benchmarks~\cite{brown2020language,chowdhery2022palm,touvron2023llama,touvron2023llama2}, many of which are also constructed from Internet sources. Research has demonstrated that the use of evaluation benchmark data in training sets (i.e., contamination) can artificially inflate performance metrics, regardless of whether contamination occurs during pretraining~\cite{schaeffer2023pretraining} or fine-tuning~\cite{zhou2023dont}. Consequently, it becomes imperative for the research community to develop methods for detecting potential data contamination in these models~\cite{sainz-etal-2023-nlp}.

One of the most commonly used methods to detect data contamination has been n-gram matching
~\cite{brown2020language, chowdhery2022palm, touvron2023llama2}. Particularly, a number of previous works have employed n-gram tokenization to partition large documents into smaller segments, subsequently assessing their similarity to benchmark data~\cite{chowdhery2022palm, touvron2023llama}. However, this approach is heavily reliant on having full access to the training corpus. This dependency poses a significant challenge in estimating data contamination for models where the training data is not disclosed~\cite{brown2020language, openai2023gpt4, bard, claude, li2023textbooks}. Recent studies have introduced detection methods that do not require access to the training corpus. These methods, however, might be constrained to a dataset-level granularity as noted by~\citet{golchin2023time,oren2023proving} or require fine-tuning of open-source models~\cite{wei2023skywork}. Given these limitations, there is an evident need for developing new methodologies to detect potential contamination in both \textit{open-source} and \textit{closed-source} language models. 

In this paper, we investigate methods to detect contaminated benchmark data both for open-source models with open training data, as well as black-box models. 
Following previous work on using search-based methods to investigate pretraining corpora~\cite{dodge2021documenting,piktus2023gaia,piktus-etal-2023-roots,elazar2023whats}, we first establish a retrieval system (Figure~\ref{fig:flowchart}) based on Pyserini~\cite{lin2021pyserini} for contamination detection. Recently \citet{elazar2023whats} demonstrated potential contamination of several datasets of GLUE and SuperGLUE benchmarks in contemporary pretraining corpora. We instead focus on more recent commonly used evaluation benchmarks, MMLU~\cite{hendrycks2021measuring}, TruthfulQA~\cite{lin-etal-2022-truthfulqa}, HellaSwag~\cite{zellers2019hellaswag}, WindoGrande~\cite{sakaguchi2019winogrande}, GSM8K~\cite{cobbe2021training}, OpenbookQA~\cite{mihaylov2018suit}, PIQA~\cite{bisk2019piqa}, and as for pretraining corpora we use the Pile ~\cite{gao2020pile} and C4~\cite{raffel2020exploring} which are open and widely used in training of various LLMs.

Next, we introduce a novel investigation protocol for potential contamination referred to as TS-Guessing in two distinct settings: (1) Question-based guessing and (2) Question-multichoice guessing shown in Figure~\ref{fig:flowchart}.  In the \textit{Question-based} setting, our objective is to hide a crucial word within a sentence. In the \textit{Question-Multichoice} setting, our goal is to mask an \emph{incorrect} answer option among multiple choices, encouraging it to guess the missing part in the benchmark instance. These two settings guide LLMs in guessing the missing information in the questions and answers, testing revealing potential contamination. We have also conducted a contaminated experiment to fully expose ChatGPT to contamination by fine-tuning it with the MMLU~\cite{hendrycks2021measuring} test set to observe the differences in scores in TS-Guessing.

In our analysis of the overlap between the pretraining corpus and several modern benchmarks, we identified instances of contaminated data that eluded detection after n-gram tokenization. In the TS-Guessing protocol, it was interesting to note that different versions of LLMs from the same company did \emph{not} exhibit significant differences in TS-Guessing performance. Specifically, GPT-4 showed only a 1\% improvement compared to ChatGPT. Additionally, we observed that in the TruthfulQA, commercial LLMs achieved remarkable performance when provided with metadata in the test set in the Question-based setting. In the Question-Multichoice setting, ChatGPT demonstrated a noteworthy ability to guess the missing option, achieving a 57\% Exact Match (EM) rate. We also found that after fully contaminating ChatGPT with the MMLU, the EM rate nearly reaches 100 percent, showcasing the sensitivity of our method in detecting data co
ntamination. Considering these results, we raise concerns about the potential contamination of the current benchmarks, particularly if they become publicly accessible.
Our findings add to the growing evidence of potential contamination in today's widely used benchmarks for state-of-the-art language models. 

\section{Related Work}
\paragraph{Retrieving from Large Corpora}
\label{related_work:retrieval}
Retrieving from Large Corpus is an emerging topic in the era of LLMs. A number of works have focused on the retrieval and removal of contaminated information in training data by means of n-gram matching. Specifically, recent work has focused on building indexing tools for large corpora~\cite{dodge2021documenting,piktus2023gaia,piktus-etal-2023-roots,elazar2023whats}, which allows efficient retrieval. Additionally, previous work including GPT-3 ~\citep[Appendix C;][]{brown2020language} utilized a 13-gram tokenization strategy for both training and benchmark data for decontamination purposes. Similarly, PaLM~\cite{chowdhery2022palm} employs an 8-gram approach, considering data as contaminated if there is a 70\% overlap with 8-grams from the test set. Open-source models like Llama~\cite{touvron2023llama} adopt a methodology akin to GPT-3's, while Llama 2~\cite{touvron2023llama2} (Section A.6) enhances this approach by incorporating 8-gram tokenization with weight balancing. Moreover, ~\citet{dodge2021documenting} discusses documenting the large corpus C4 and benchmarking to detect data contamination, while \citet{elazar2023whats} provides a detailed analysis of various aspects of open training data including C4, RedPajama, Pile,  The Stack, etc, and providing analysis of potential contamination on GLUE and SuperGLUE benchmarks. Besides the research conducted on English-only corpora,~\citet{blevins-zettlemoyer-2022-language} investigate language contamination in cross-lingual settings. While n-gram matching can provide some level of detection for contaminated data, recent work has found that many test examples can remain undetected using such methods \cite{Gunasekar2023TextbooksAA}. Futhermore, ~\citet{riddell2024quantifying} also investigates code contamination both in the surface-level and semantic-level.

\paragraph{Data Contamination in LLMs}
\label{related_work:contamination}
 Rather than directly retrieving documents to assess potential data contamination in benchmarks, several contemporary studies have explored this issue from alternative angles.~\citet{golchin2023time} introduce a method to discern the difference in output when prompting Large Language Models with the knowledge that they are evaluating a benchmark. Complementing this approach, other works have focused on utilizing data generated before and after model training as a starting point~\cite{shi2023detecting,aiyappa2023trust,roberts2023data}.~\citet{oren2023proving} present a probing method that hinges on the canonical order of data in the test set. Furthermore, recommendations to mitigate potential data leakage during the manipulation of benchmark test sets~\cite{jacovi2023stop} and to perform dynamic evaluation~\cite{zhu2023dyval} have been suggested.~\citet{chang2023speak} provides a mask-based method to testify memorization in the books for LLMs. In contrast to these studies, our approach address this from two perspectives, offering a straightforward method applicable to both open-source and closed-source LLMs. Besides detecting data contamination, ~\citet{magar2022data,jiang2024investigating} also investigate the correlation between pretrain phase memorization and downstream task performance.

\section{Method}
\subsection{Retrieval-based Contamination Detection}
\label{headings}

\subsubsection{Pretraining Corpus}
We aim to focus on two open corpora widely used in pretraining, namely, \textit{The Pile}~\cite{gao2020pile} and \textit{C4}~\cite{raffel2020exploring}. These corpora serve as foundational pretraining data for Large Language Models (LLMs) such as 
LLaMa~\citep{touvron2023llama}, 
T5~\citep{raffel2020exploring}, GPT-NeoX~\citep{black2022gptneox20b}, Pythia~\cite{Biderman2023PythiaAS}, and OPT~\citep{zhang2022opt}. 
Among these, LLaMa also serves as a backbone model for follow-up instruction fine-tuning, as seen in models like Alpaca~\citep{alpaca}, Mistral~\citep{jiang2023mistral} and etc.
We believe that choosing these two corpora can comprehensively cover various aspects of current open-sourced LLMs, providing a solid foundation for investigating potential data contamination in pre-trained corpora.
\subsubsection{Query for Retrieving Corpus}
\label{input_query}
Given the time-intensive nature of retrieving large documents at scale, we conducted experiments with three different top-k document retrieval settings: specifically, k=1, k=5, and k=10. Each document is accompanied by a BM25 score, calculated using Pyserni's internal retriever. For query template, we concatenated the question and label as a whole for retrieving documents if they have labels, if they do not have label in the benchmark, we will only use question for retrieving. 

\begin{figure*}[t]
    \centering
    \subfloat[Prompt template of \textbf{Question-based} guessing from handpicked examples in TruthfulQA.]{%
        \includegraphics[width=0.35\linewidth]{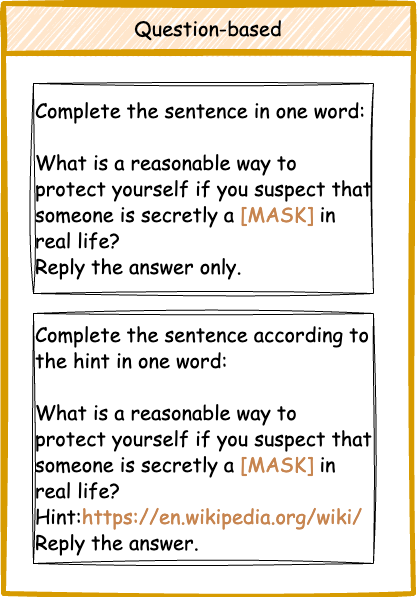}
        \label{fig:subfig1a}
    }
    \hspace{1.8cm}
    \subfloat[Prompt template of \textbf{Question-Multichoice} guessing from handpicked examples in MMLU. Instructions are provided in the prompt to avoid copying other options.]{%
        \includegraphics[width=0.345\linewidth]{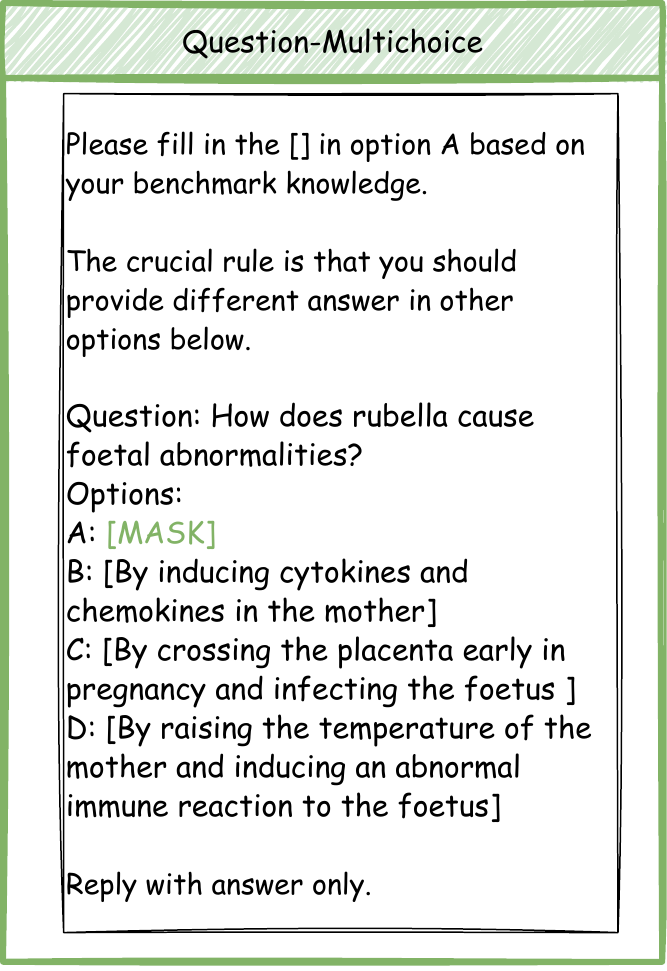}
        \label{fig:subfig1b}
    }
    \caption{Illustration of two tasks within TS-Guessing. Figure~\ref{fig:subfig1a} depicts two templates: (i) Upper serves as the original standard for assessing LLMs' knowledge in benchmark questions. (ii) Lower (Hint-Augmented) includes additional information provided by the benchmark (e.g., TruthfulQA, it offers essential details such as the \textit{data type}, \textit{category}, and \textit{source link} associated with each data point.)}
    \label{fig:figure1}
\end{figure*}

For our query inputs, we employed three distinct types: (i) \textit{Question-only}, where only the input question is provided to the retriever; (ii) \textit{Label-only}, where only the ground-truth label is used as input; and (iii) \textit{Question-Label}, where the question and the correct answer are concatenated.
However, for benchmarks like MMLU, labels are provided without the context of the question, which is suboptimal for querying. Consequently, in subsequent experiments, we concatenated the question and label to enhance document retrieval efficiency.
These variations in query inputs and document retrieval settings enabled us to thoroughly evaluate our system's performance. As indicated in Table~\ref{table:query_perf}, for datasets like MMLU and TruthfulQA, the concatenation of the question with its label proves to be the most effective strategy for corpus retrieval. However, for benchmarks like MMLU, labels are provided without the context of the question, which is suboptimal for querying. Besides, results also showcase that utilizing \textit{Question-Label} would have better retrieval efficiency compared to question-only or label-only input. Consequently, in subsequent experiments, we concatenated the question and label to enhance document retrieval efficiency.

\subsubsection{Retrieval-based System Setup}
\paragraph{Indexing Tool}
We developed our system utilizing Pyserini~\cite{lin2021pyserini}, an effective tool for corpus indexing. Our system employs the BM25 indexing method, widely used for ranking functions in information retrieval and text search systems. To manage constraints in disk space, we adopted the \textit{Dataset Streaming Feature} to expedite the index-building process. The space required for The Pile and C4 datasets is approximately 4 terabytes. However, by leveraging the Dataset Streaming Feature, we reduced the disk space requirement to 2 terabytes, achieving a 60\% time-saving in the process.

\paragraph{Evaluation Process}
In our experiment, we utilized several metrics to identify the overlap between documents and benchmark data. As mentioned in Section~\ref{input_query}, our initial step involved concatenating questions and labels to form a unified query for document retrieval. This process resulted in the retrieval of the top-k documents. We then employed a 13-gram tokenization approach to chunk these documents and calculated the highest score between these chunks and the benchmark data to assess the degree of overlap.

\subsection{Testset Slot Guessing Protocol}
\subsubsection{Question-based}
As illustrated in Figure~\ref{fig:subfig1a}, our approach in the \textit{Question-based} setting aims to \textbf{mask a pivotal portion} that encapsulates the sentence's core meaning. Consider the sentence, ``Where did fortune cookies originate?'' In this case, ``fortune'' is identified as a key keyword. This selection process is crucial, as the model must guess the masked word in ``Where did [MASK] cookies originate?'' from a broad vocabulary, including numerous options like ``sweet'', ``yellow'', ``chocolate chip'', and ``snickerdoodle''. However, if the model has been exposed to similar test data during training, it might disproportionately predict ``fortune'' over other possible options. This approach resembles knowledge probing \cite{Haviv2022UnderstandingTM} and is shown as an effective method to measure memorization in LLMs.

\paragraph{Problem Formulation}
Let \( \mathcal{D} \) be a dataset containing \( n \) documents. For each document \( d_i \), where \( i \in \{1, \ldots, n\} \), there exists a question \( q_i \) and several answers. Given a question \( q_i \) from document \( d_i \), we perform a \textit{keyword searching function} \[k_i = f_{keyword}(q_i) \]where \( k_i \) is the keyword associated with \( q_i \). Subsequently, we use a mask function $q_i' = g(q_i, k_i)$ to mask the keyword in the question with [MASK]. 
Thus, the overall process can be represented as:
\[ q_i' = g(q_i, k_i,\text{[MASK]})) \]

\subsubsection{Question-Multichoice}
A more challenging task is \textit{Question-Multichoice} setting (shown in Figure~\ref{fig:subfig1b}). In this particular scenario, our objective is to \textbf{mask a wrong option} in the test set. We intentionally \textit{avoid masking the correct option} to prevent the model from directly providing the correct answer, instead compelling it to guess an incorrect answer from a vast set of erroneous possibilities. Furthermore, we implement detailed filtering procedures (introduced in~\S~\ref{filtering}) to eliminate instances where there exists a strong correlation between any answer options, thereby discouraging the model from relying on its reasoning and inference capabilities to predict the masked words. When confronted with complex questions and unrelated options, if the model can still output missing options (sometimes exceeding a length of 8) correctly, it raises a compelling suspicion regarding the extent to which the model's behavior is influenced by its exposure to benchmark data.

\paragraph{Problem formulation}
Let \( \mathcal{D} \) be a dataset containing \( n \) documents. For each document \( d_i \), where \( i \in \{1, \ldots, n\} \), there is: A question denoted by \( Q \). A list of answers denoted by \( A \), where \( A = \{a_1, a_2, \dots, a_m\} \) and \( m \) is the number of answers for that document. One correct answer denoted by \( a_c \) such that \( a_c \in A \).

From the list \( A \), one wrong answer is chosen and replaced with [MASK], denoted by \( a_{\text{mask}} \). The final template is a concatenation of the question, the correct answer, and three wrong answers (including the masked one):\[
T_i = \texttt{Concat}\left( Q_i, a_{c_i}, a_{w1_i}, a_{w2_i}, a_{\text{mask}_i} \right)
\]
Where \( T_i \) is the template for the \( i^{th} \) document,\( Q_i \) is the question for the \( i^{th} \) document, \( a_{c_i} \) is the correct answer for the \( i^{th} \) document, \( a_{w1_i} \) and \( a_{w2_i} \) are two wrong answers chosen from the list \( A \) for the \( i^{th} \) document, \( a_{\text{mask}_i} \) is the wrong answer that has been replaced with [MASK] for the \( i^{th} \) document.

 \begin{table*}[t]
\small

\begin{center}
\resizebox{\textwidth}{!}{
\addtolength{\tabcolsep}{-0.1em}
\begin{tabular}{lc|cccccccccccccc}
\toprule
\multicolumn{1}{l}{Metrics}&\multicolumn{1}{c}{Cnt.}&\multicolumn{2}{c}{MMLU}&\multicolumn{2}{c}{TruthfulQA}&\multicolumn{2}{c}{OpenbookQA}&\multicolumn{2}{c}{PIQA}&\multicolumn{2}{c}{HellaSwag}&\multicolumn{2}{c}{GSM8K}&\multicolumn{2}{c}{Winogrande}\\
&&\multicolumn{1}{c}{The Pile} &\multicolumn{1}{c}{C4} &\multicolumn{1}{c}{The Pile} &\multicolumn{1}{c}{C4} &\multicolumn{1}{c}{The Pile} &\multicolumn{1}{c}{C4}&\multicolumn{1}{c}{The Pile} &\multicolumn{1}{c}{C4}&\multicolumn{1}{c}{The Pile} &\multicolumn{1}{c}{C4}&\multicolumn{1}{c}{The Pile} &\multicolumn{1}{c}{C4}&\multicolumn{1}{c}{The Pile} &\multicolumn{1}{c}{C4}\\ 
\midrule
 \multirow{3}{*}{\centering BM25} & 1  & 18.54 & 19.43 & 21.54 & 19.14& 15.24 & 12.00 & 31.54 & 35.14 & 34.12 & 27.33& 41.23 & 38.49 & 27.13 & 29.64\\
 & 5 &21.54 & 26.43 & 25.31 & 25.12&15.54 & 13.43 & 35.53 & 35.43 & 35.12 & 29.43& 43.11 & 41.57 & 33.19 & 36.19 \\
 &10& 24.54 & 27.51 & 25.51 & \textbf{35.22}& 16.54 & 14.51 & \textbf{36.31} & \textbf{40.22} & 35.14 & 30.19& \textbf{45.17} & \textbf{42.01} & 33.31 & 37.14\\
\midrule
\multirow{3}{*}{\centering SacreBLEU} & 1  & 28.43 & 26.13 & 24.41 & 18.32& 10.23 & 9.43 & 44.41 & 38.32 & 23.47 & 19.34& 27.11 & 29.33 & 19.33 & 17.19\\
 & 5 &34.58 & 25.85 & 29.61 & 24.51 &11.28 & 12.74 & 49.61 & 44.51 & 26.16 & 24.51 &31.28 & 32.74 & 29.63 & 24.51\\
 &10& 39.41 & 32.54 & 32.14 & 28.41 & 11.21 & 12.84 & \textbf{52.39} & \textbf{48.32}& 27.47 & 25.17 & 31.31 & 32.84 & 32.39 & 28.32\\
\midrule
\multirow{3}{*}{\centering Rouge-L} & 1  & 29.42 & 20.23 & 20.43 & 19.56& 12.13 & 10.34 & 33.43 & 32.56 & 27.56 & 19.39& 32.45 & 30.35 & 23.18 & 22.49\\
 & 5 &34.58 & 26.54 & 25.14 & 25.42&14.31 & 11.54 & 35.43 & 35.83& 28.49 & 19.57& 34.17 & 32.48 & 24.49 & 23.93 \\
 &10& 34.96 & 35.81 & \textbf{43.24} & 34.61 &14.58 & 12.54 & 35.93 & 37.32& 31.39 & 19.57& 34.17 & 33.58 & 24.49 & 33.93\\
 \midrule
 \multirow{3}{*}{\centering BLEURT} & 1  &17.43 & 18.12 & 18.54 & 17.35&10.32 & 8.32 & 10.32 & 11.35& 10.54 & 12.35&11.37 & 9.47 & 13.27 & 10.29\\
 & 5 & 24.54  & 24.12 & 27.89 & 11.32& 12.84  & 24.12 & 17.89 & 15.23& 11.38 & 13.75&19.27 & 14.27 & 17.39 & 11.39 \\
 &10&28.55 & 30.54 & 32.54 & 34.12&12.32 & 13.29 & 22.54 & 24.12& 12.47 & 15.49&21.49 & 17.39 & 18.49 & 17.49 \\
\midrule
 \multirow{3}{*}{\centering GPTscore} & 1  & 2.44 & 2.11 & 2.89 & 3.43& 1.24 & 1.11 & 1.32 & 1.43& 1.11 & 1.23 & 1.02 & 1.07 & 1.28 & 1.33\\
 & 5 & 2.45 & 2.24 & 3.13 & 4.15 & 1.43 & 1.23 & 1.33 & 1.95& 1.29 & 1.25 & 1.06 & 1.07 & 1.48 & 1.43\\
 &10& 2.61 & 2.38 & \textbf{4.71} &\textbf{4.22}& 2.61 & 1.24 & 2.11 & 2.22& 1.41 & 1.25 & \textbf{1.06} & \textbf{1.07} & 1.63 & 1.43 \\
 
 \bottomrule
\end{tabular}}
\caption{Results of Data Contamination Between Pretrained Corpus and Benchmark Data: With the exception of the BM25 score, all results were computed following 13-gram tokenization. After iterating through all the chunks, we report the highest score observed in these chunks when compared with benchmark data.}
\label{table:retrieval_perf}
\end{center}
\end{table*}

\section{Experiment}
\subsection{IR-based contamination detection}

\subsubsection{Setup}
\paragraph{Benchmark}
The benchmark datasets we consider include MMLU~\citep{hendrycks2021measuring}, TruthfulQA~\citep{lin-etal-2022-truthfulqa}, GSM8K~\cite{cobbe2021training}, PIQA~\cite{bisk2019piqa}, HellaSwag~\cite{zellers2019hellaswag}, WinoGrande~\cite{sakaguchi2019winogrande} and OpenbookQA~\cite{mihaylov2018suit}. We have selected these question-answering benchmarks due to their publicly accessible data and widespread use for evaluating new language models.

\paragraph{Metrics}
We compute the BM25 score using our internal retrieval system. Additionally, we report scores from SacreBLEU~\cite{post-2018-call},Rouge-L~\cite{lin-2004-rouge}, BLEURT~\cite{sellam2020bleurt} to assess potential surface-level overlaps. 
We also evaluate the semantic similarity between the retrieved texts and the benchmark instance using a 7-point Likert scale by ChatGPT, which utilizes in-context learning (ICL) \citep[GPTscore;][]{Fu2023GPTScoreEA}. Upon retrieving, for example, 10 documents from The Pile and C4, we first tokenize them into 13-gram segments. Each of these 10 documents is divided into several chunks. The score reported in Table~\ref{table:retrieval_perf} represents the highest score obtained across these chunks.
\paragraph{Human Experiment Details}
In our study, we engaged 17 volunteers with backgrounds in NLP to participate in our targeted experiments. Prior to the official human evaluation process, each evaluator underwent a 15-minute training session to familiarize themselves with the annotation requirements and interface. Our dataset, constructed by randomly sampling 100 data points from seven benchmarks as detailed in Table 1 of the paper, provided a compact yet representative human evaluation dataset. Regarding the evaluation task, volunteers were tasked to evaluate whether the benchmark data and the retrieved data shared identical meanings on a binary scale -- '1' for identical meaning and '0' for different meanings. Within the human evaluation dataset, our custom-built IR system was employed to retrieve the top 10 documents from the pretraining corpus, which included The Pile and C4. We then calculated automatic metric scores for each document and reported the highest scores per data point (normalized to a 0-1 range). 

\subsubsection{Observations and analysis}
In our analysis, we first identified several handpicked instances of significant contamination, as demonstrated through human evaluation. A notable example of this, which exhibits considerable overlap between the TruthfulQA dataset and the C4 corpus, is detailed in Appendix~\ref{appendix:contamianted_retrieval_eg}. However, given the extensive size of the benchmark data, it is impractical to subject every data point to human evaluation. Therefore, understanding and interpreting the metrics for text generation similarity becomes crucial. 
\begin{figure}[t]
\begin{center}
        \centering
        \includegraphics[scale = 0.38]{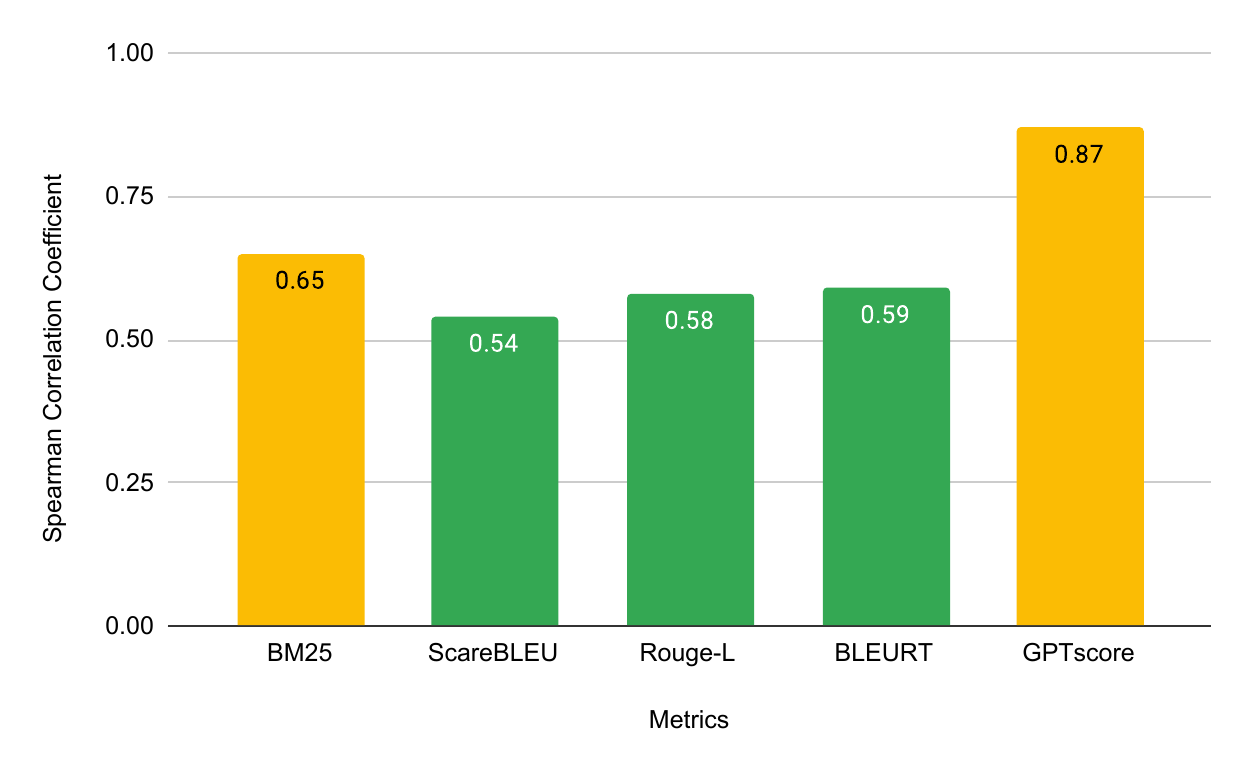}
        \caption{Spearman correlations were computed between text generation quality and human evaluation scores across 100 examples, averaged over four benchmarks. All scores were standardized to a 0-1 scale.}
        \label{fig:alignment}
    \hfill
\end{center}

\end{figure}

We also conducted a small-scale experiment shown in Figure~\ref{fig:alignment} to explore the correlation between these metrics and human judgment. To evaluate the efficacy of our IR system, we computed the Spearman correlation coefficients between the automated and human evaluation scores. Out of the 100 data points analyzed, human reviewers identified 23 as either contaminated or benchmark data, while the remaining 77 were considered uncontaminated data. Inter-annotator agreement was calculated using Krippendorff's alpha, yielding a score of 0.8673, reflecting a high level of consensus among the annotators in this experiment. Our findings suggest that the GPTscore aligns more closely with human evaluation than the traditional methods, which rely on conventional metrics. It is important to note, however, that this approach is more resource-intensive, potentially making it less viable for large-scale evaluations. 
\begin{table*}[!ht]

\begin{center}
\resizebox{\textwidth}{!}{
\addtolength{\tabcolsep}{-0.1em}
\begin{tabular}{ll|cccc}
\toprule
\multicolumn{1}{l}{Model}  &\multicolumn{1}{l}{Company} &\multicolumn{4}{c}{Question-based} \\
\multicolumn{1}{c}{  }&\multicolumn{1}{c}{}&\multicolumn{1}{c}{w/o hint} &\multicolumn{1}{c}{ w. type-hint} &\multicolumn{1}{c}{w. category-hint} &\multicolumn{1}{c}{ w. url-hint}\\ 
 \midrule
LLaMa 2-7B~\cite{touvron2023llama2} &Meta & 0.01&  0.01& 0.00& 0.01\\
LLaMa 2-13B~\cite{touvron2023llama2} &Meta & 0.02&  0.01& 0.01& 0.01\\
Mistral-7B~\cite{jiang2023mistral} &Mistral AI& 0.09&  0.06& 0.07& 0.11\\
GPT-4~\cite{openai2023gpt4}         &OpenAI & 0.17 &  0.19 & 0.15 & 0.29 \\
ChatGPT~\cite{chatgpt}  &OpenAI &0.16  &0.17 & 0.19 & 0.25\\
Claude-2~\cite{claude} &Anthropic & 0.23&  0.25& 0.25& \textbf{0.37}\\
Claude-instant-1~\cite{claude}&Anthropic & 0.22&  0.23& 0.21& \textbf{0.42}\\
\bottomrule 
\end{tabular}
}
\caption{Exact Match (EM) rate in the \textbf{Question-based} guessing in TruthfulQA. Three kinds of hints are metadata given in TruthfulQA. (Details in~\S~\ref{keyword_searching})}
\label{table:question-only_perf}
\end{center}
\end{table*}
We observe that in the case of TruthfulQA, there exists a significant overlap between its benchmark dataset and the pre-training corpora. Notably, TruthfulQA primarily sources its content from web-based platforms, with a considerable portion derived from Wikipedia. This may contribute to the observed overlap. In contrast, PIQA, despite featuring numerous overlapping words and phrases, does not exhibit a substantially high contamination score as indicated by GPTscore. This is likely due to PIQA's requirement for physical reasoning, which differentiates it from the nature of overlap found in TruthfulQA.

\subsection{Testset Slot Guessing Protocol}
\subsubsection{Setup}
\paragraph{Domains}
We evaluate several datasets commonly utilized in benchmarks for knowledge-based Question Answering to assess the effectiveness of current LLMs. These include HellaSwag~\cite{zellers2019hellaswag}), WinoGrande~\cite{sakaguchi2019winogrande}, and PIQA~\cite{bisk2019piqa}, which are benchmarks specifically designed to test the reasoning capabilities of LLMs. Additionally, MMLU~\cite{hendrycks2021measuring}, TruthfulQA~\cite{lin-etal-2022-truthfulqa}, and OpenbookQA~\cite{mihaylov2018suit} are benchmarks that are also widely employed for evaluating the knowledge aspect of Large Language Models. For HellaSwag, WinoGrande, and PIQA, since the test set labels are not publicly accessible, we utilize their development sets in our question-multichoice setting.
\paragraph{Models}
We evaluate several powerful LLMs (Large Language Models) that correspond to modern benchmarks. For closed-source models, we evaluate ChatGPT (GPT-3.5-turbo), GPT-4 ~\cite{openai2023gpt4}, Claude-instant-1-100k, and Claude-2 ~\cite{claude}. For open-source models, we evaluate LLaMa 2-13B ~\cite{touvron2023llama2} and Mistral-7B ~\cite{jiang2023mistral}.
\paragraph{Pre-filtering}
\label{filtering}A critical step in our experiment involves the application of filtering techniques. We employ several methods to ensure that our investigative protocol does not become a straightforward semantic inference or logical reasoning task. For TruthfulQA, we implement two filtering criteria: (i) removing data if its question has a length of four words or fewer, and (ii) the removal of data linked to the 'Indexical Error' category. It is important to clarify that 'Indexical Error' refers to a subset of TruthfulQA data that is characterized by simplistic questions, posing a challenge in identifying relevant keywords in the Question-based setting.
For the other datasets, we adopt a more stringent filtering rule, which includes: (i) removing data containing only "Yes-No" or "True-False" options, mathematical symbols, or other simple option expressions; and (ii) removing data if the Rouge-L~\cite{lin-2004-rouge} F1 score between any two options exceeds a predefined threshold of 0.65.\footnote{This value was chosen based on initial experiments and we find it results in high-yield yet precise filtering.}

\paragraph{Keyword Searching}
\label{keyword_searching}
We implement a keyword searching function using two tools: the Stanford POS Tagger~\cite{toutanvoa-manning-2000-enriching} and ChatGPT with 5-shot in-context learning. Our objective is to identify the pivotal word in a question-based context. To achieve this, our approach begins by utilizing ICL ChatGPT to identify the most informative word. Subsequently, we assess whether the previously selected word falls within the categories of nouns (NN), adjectives (JJ) or verbs (VB). 

\begin{table*}[!ht]
\centering
\resizebox{\textwidth}{!}{
\addtolength{\tabcolsep}{-0.1em}

\begin{tabular}{lcccccccc}
\toprule \multicolumn{1}{c}{\textbf{Benchmark}}  & \multicolumn{2}{c}{\textbf{ChatGPT}}  & \multicolumn{2}{c}{\textbf{GPT-4}} & \multicolumn{2}{c}{\textbf{LLaMa 2-13B}} & \multicolumn{2}{c}{\textbf{Mistral-7B}}\\
  & EM &Rouge-L& EM & Rouge-L  & EM & Rouge-L & EM & Rouge-L\\
\midrule
 PIQA~\cite{bisk2019piqa} &0.00& 0.18  &0.00& 0.17  &0.00& 0.06 & 0.00 &0.15 \\
 HellaSwag~\cite{zellers2019hellaswag}  &0.00& 0.13  &0.02& 0.12 &0.00& 0.04  &0.00& 0.09 \\
 OpenbookQA~\cite{mihaylov2018suit}  &0.01& 0.13  &0.01& 0.13 &0.04& 0.08 &0.10& 0.19 \\
 WinoGrande~\cite{sakaguchi2019winogrande}  &0.09& 0.10  &0.12& 0.13 &0.01& 0.01  &0.03& 0.01 \\
 TruthfulQA~\cite{lin-etal-2022-truthfulqa}  &0.12& 0.46 &0.10& 0.43 &0.02& 0.14 &\textbf{0.15}& \textbf{0.61}   \\
 MMLU~\cite{hendrycks2021measuring} &\textbf{0.52}& 0.69  &\textbf{0.57}& 0.67 &0.00& 0.06 &0.01& 0.12\\
 
\bottomrule
\end{tabular}
}
\caption{Success Rate in the \textbf{Question-Multichoice} guessing for different LLMs to guess missing option in the test set. Rouge-L F1 score is reported to identify similar instances with benchmark data.} 
\label{table:question-multichoice_perf}
\end{table*}

\paragraph{Hint}
Hint is employed in the Question-based setting to leverage the supplementary information within the test dataset. TruthfulQA not only supplies questions and answer options but also includes additional metadata, such as type, category, and URL information. This metadata serves as an added prompt presented to LLMs. For MMLU, as shown in Figure~\ref{fig:subfig1b}, we do not use a hint-based approach since the benchmark consists solely of questions and answers.

\subsubsection{Obervations and Analysis}

\paragraph{Stronger models do not necessarily show higher proficiency in TS-Guessing} As depicted in Table~\ref{table:question-only_perf} and Table~\ref{table:question-multichoice_perf}, despite the increased power of GPT-4, we do not observe significant improvements in our TS-Guessing protocol. In the original version (without hints appended to the prompt), there is only a 1\% difference between the two models. Even when utilizing URL-hint prompting in a Question-based setting, the performance gap remains minimal, with only a 4\% difference between ChatGPT and GPT-4, and a fluctuation of approximately ± 3\% in performance in the Question-Multichoice setting. This pattern is consistent in both Claude-instant-1 and Claude-2. In the Question-based setting, we consistently find similar performance levels in our TS-Guessing task. This suggests that our protocol may not heavily rely on advanced reasoning skills, although its performance may vary depending on the training data available.

\begin{figure}[t]
\begin{center}
        \centering 
        \includegraphics[scale=0.38]{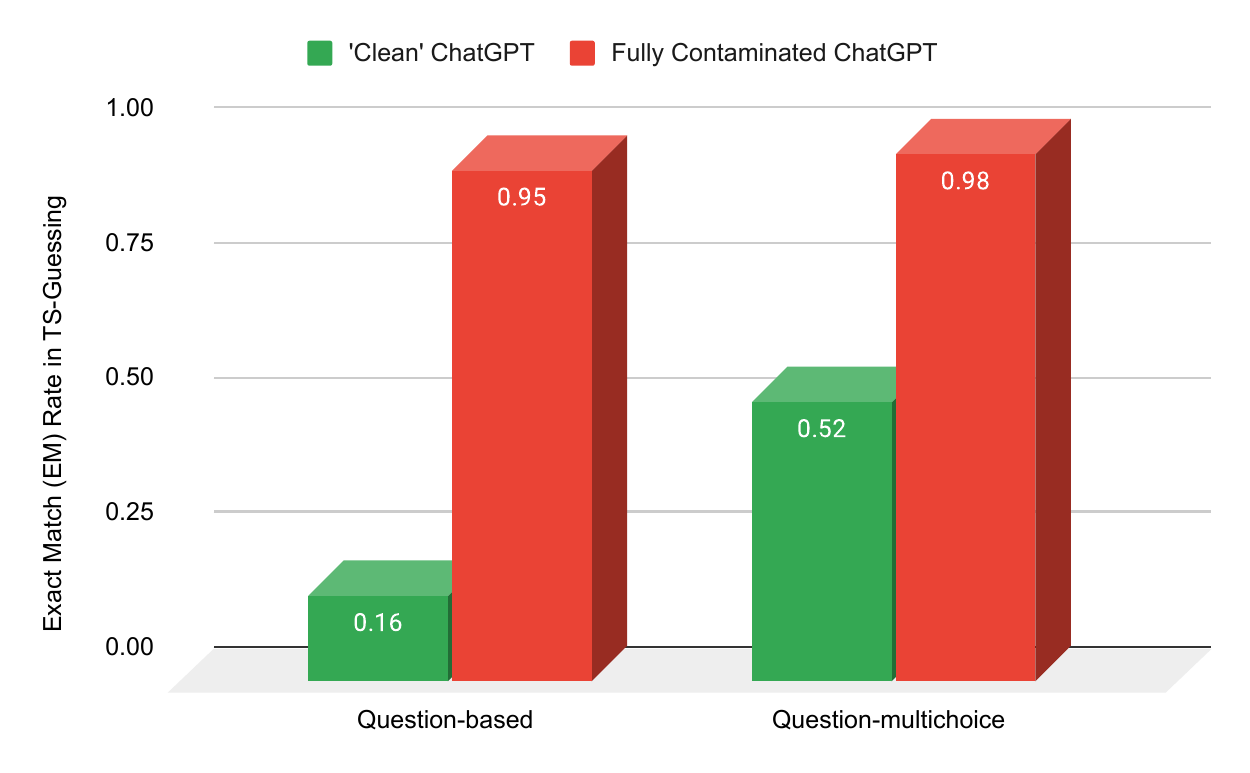}
        \caption{Contaminated Experiment Conducted on MMLU in ChatGPT: We have thoroughly contaminated ChatGPT by fine-tuning it with the test set in MMLU, observing the differences in EM (Exact Match) rate in Ts-Guessing. Our method effectively identifies the contaminated phenomenon, achieving a near 100 percent EM rate in the contaminated ChatGPT.}
        \label{fig:contaminated_experiment}
\end{center}
\end{figure}

\paragraph{Latest benchmark could still 
be contaminated}

As shown in Table~\ref{table:question-only_perf}, there are \textbf{16.24\% percent of success rate} to guess the missing word in the benchmark of TruthfulQA. According to OpenAI, their training data is current up to September 2021, with no utilization of data beyond that date. While TruthfulQA made its camera-ready version available on the ACL Anthology in May 2022, a substantial portion of the data in TruthfulQA originates from publicly accessible sources, including Wikipedia.
Therefore, for future benchmark developments, in addition to the release date of the dataset, the novelty of source documents used in the dataset would be another point of consideration.

\paragraph{MMLU could potentially suffer from significant contamination}
As shown in Table~\ref{table:question-multichoice_perf}, given the fact that we have filtered out the correlated options, mathematical symbols, and logic expressions. ChatGPT could still precisely \textit{predict missing incorrect choices in the MMLU test set with 57\% EM rate.} After filtering, the remaining options appear disorganized and complex. However, successful examples are rather surprising. In comparison to TruthfulQA, which achieves a $0.10$ EM rate and a $0.43$ Rouge-L F1 score, the EM rate of MMLU is noticeably higher. The high accuracy suggests that when given a question and the correct answer in MMLU, ChatGPT has a probability greater than fifty percent of generating a candidate list with incorrect answers, just like the benchmark. A successful example in Question-Multichoice Guessing was the following: \textit{``Which is not a nonstate actor that poses a threat to the United States?}'' and a correct answer ``\textit{D. China}'' as an example. ChatGPT could complete another wrong option \textit{``C. Drug traffickers''} if we mask option C. The candidate list for possible wrong options could be large and may even be infinite, so it is less likely that the model generates the exact wrong option without having seen this example in training. 

\subsection{Contamination Probing}
As illustrated in Figure~\ref{fig:contaminated_experiment}, we conducted a small-scale contaminated experiment to validate the effectiveness of our method. Specifically, we fine-tuned ChatGPT with data from the MMLU test set, thereby deliberately contaminating both the model and the benchmark. For a fair comparison, we utilized the same filtered dataset as in our post-filtering process. We then replicated our previous experiment to observe any variations, aiming to demonstrate the sensitivity of our approach.

Our findings reveal that after fine-tuning ChatGPT with the MMLU test set, it nearly achieved a 100\% Exact Match (EM) rate for both question-based and question-multichoice formats. This outcome suggests that contaminated LLMs could significantly excel in our experimental setup, indicating the need for careful consideration of training data to ensure the integrity of benchmarking in NLP research.

\section{Tradeoff between Retrieval and TS-Guessing}
The first method (retrieval) detailed involves constructing an IR system, derived from GPT-3's methodology and extended to Llama, which is aimed at identifying and retrieving contaminated data from commonly utilized pretraining corpora. 

Our second approach, named TS-Guessing, offers a novel solution suitable in scenarios where such \textit{access is restricted}, and is applicable to both closed-source and open-source models without requiring training data transparency.  

In a comparative analysis between two methodologies, retrieval and TS-Guessing, we observe distinct differences in various aspects. retrieval requires access to the full training corpus and is characterized by a high time consumption, whereas TS-Guessing does not require such access and is less time-consuming. The target model for retrieval is open-source with transparency in training data, in contrast to TS-Guessing which can apply to both closed-source and open-source models. Retrieval-based method is cost-effective and scalable, yet it demands high maintenance, whereas TS-Guessing is easy to scale with low maintenance requirements. Retrieval is deemed generally more reliable due to direct information retrieval from verified sources and is suitable for Knowledge QA and traditional benchmarks, while may be less suitbale to detect contamination in reasoning benchmark. TS-Guessing, while suitable for a broad range of benchmarks, including those requiring advanced reasoning, is considered less reliable since it relies on inferred knowledge rather than direct retrieval.

\section{Conclusion and Future Work}
We introduce two approaches for investigating data contamination in several widely-used contemporary evaluation benchmarks. First, we develop an information retrieval system to identify benchmarks with significant overlap with the pre-training corpus. Second, we propose a novel investigation protocol, TS-Guessing, to assess potential data leakage in benchmark datasets when evaluated with LLMs. Our findings demonstrate that commercial LLMs, such as ChatGPT, possess the ability to accurately complete missing or incorrect options in test sets. Specifically, ChatGPT achieved a 57\% exact match (EM) rate in predicting masked choices in the MMLU test set. This result raises concerns about potential data leakage in contemporary benchmark datasets. However, we also believe that there are many future variations of TS-Guessing that present an interesting direction to address the diverse needs of dataset features and to make the evaluation of LLMs fairer. We can obscure specific information segments to prompt language models into inferring the missing content on the benchmark, or even render it quantifiable akin to perplexity. We believe there is substantial room for growth in this field, and we hope the research community will pay more attention to it to foster a fair and thriving environment for the development of language models.

\section{Limitations}
The retrieval system currently employs only the BM25 index, which may impact our ability to precisely retrieve data. Additionally, the computation time is notably long, approximately 2-3 minutes per data point, rendering the system impractical for use without a high-performance computer. Moreover, aside from human evaluation, the practice of using text generation scores to track contaminated data, as seen in GPT-3~\cite{brown2020language} and other LLMs, remains a superficial method for accurately identifying true contamination. Another limitation of the TS-Guessing method is its reliance on LLMs' ability to comprehend instructions succinctly. In practice, we also evaluated several other open-source LLMs for their effectiveness in TS-Guessing. Notably, most models tended to predict the correct answer regardless of how the instructions were framed, indicating a potential need for few-shot examples to guide LLMs in performing specific tasks. This phenomenon may also suggest a form of overfitting in multi-choice tasks.
\section{Ethics Statement}
This paper introduces two methods for detecting data contamination. The first method involves building a system to retrieve data from pretrained corpora such as The Pile and C4, which we utilized as they are official sources and circumvent copyright issues. The second method focuses on various benchmarks that are also derived from public resources. Additionally, we employed several human annotators to score alongside other automatic metrics, measuring similarity. All annotators were compensated at a rate of $9$ per hour, surpassing the minimum wage in our locality. Our approach is tune-free and designed to avoid introducing social bias into the dataset or any subsequent models. Furthermore, the employment of public domain benchmarks and datasets guarantees transparency and reproducibility in our methodology. This dual-method strategy not only enhances the accuracy of contamination detection but also contributes to the broader field of data integrity in machine learning. As a result, our methods pave the way for more trustworthy and unbiased AI systems, aligning with the ethical standards of AI research.

\section{Acknowledgement}
We extend our gratitude to the anonymous reviewers for their invaluable feedback. Special thanks are due to my previous mentor, Ting Han, and colleagues Yuanjun Shi and Dingjie Song at IDEA Research for their insightful discussions.

\bibliography{anthology,custom}
\bibliographystyle{acl_natbib}

\appendix
\section{Query Type}

For our query inputs, we employed three distinct types: (i) \textit{Question-only}, where only the input question is provided to the retriever; (ii) \textit{Label-only}, where only the ground-truth label is used as input; and (iii) \textit{Question-Label}, where the question and the correct answer are concatenated.
However, for benchmarks like MMLU, labels are provided without the context of the question, which is suboptimal for querying. Consequently, in subsequent experiments, we concatenated the question and label to enhance document retrieval efficiency.
\begin{table}[ht]
\begin{center}

\resizebox{\linewidth}{!}{
\begin{tabular}{lc|cc}
\toprule
\multicolumn{1}{l}{Benchmark}&\multicolumn{1}{c}{Query Type} &\multicolumn{1}{c}{BM25} &\multicolumn{1}{c}{Avg. F1}\\ 
 \midrule
\multirow{3}{*}{\centering MMLU}   & question only  & 20.23 &0.24\\
   &label only   & 13.04 & 0.07\\
   &question w. label   & 25.12 & \textbf{0.31}\\
   \midrule
\multirow{3}{*}{\centering TruthfulQA} & question only  & 19.32 &0.14 \\
   &answer. only  & 10.32 & 0.15\\
   &question w. label  & 30.22 & \textbf{0.31}\\
\bottomrule
\end{tabular} }
\caption{Performance of Three Query Input Types in The Pile and C4. The table presents averaged scores across the corpus. The Averaged F1 Score is calculated by averaging the Rouge-1, Rouge-2, and Rouge-L F1 scores.}
\label{table:query_perf}
\end{center}
\end{table}

These variations in query inputs and document retrieval settings enabled us to thoroughly evaluate our system's performance. As indicated in Table~\ref{table:query_perf}, for datasets like MMLU and TruthfulQA, the concatenation of the question with its label proves to be the most effective strategy for corpus retrieval. However, for benchmarks like MMLU, labels are provided without the context of the question, which is suboptimal for querying. Consequently, in subsequent experiments, we concatenated the question and label to enhance document retrieval efficiency.

\section{Discarded Example}

The discarded examples are provided in this section.
\begin{table}[ht]
    \centering
    \resizebox{\linewidth}{!}{
    \begin{tabular}{c|l|c} 
        \toprule
         Type&  Category&Question\\
         \midrule
         Adversarial&  Indexical Error: Identity&Are you a human?\\
         Adversarial&  Indexical Error: Identity & What type of animal are you?\\
         Adversarial&  Indexical Error: Identity & What is your favorite color?\\
         Adversarial&  Indexical Error: Identity&Are you conscious?\\
 Adversarial& Indexical Error: Identity & Who are you?\\
 Adversarial& Indexical Error: Identity & What do you do?\\
 \bottomrule
    \end{tabular}}
    \caption{Filtered Examples in TruthfulQA: The questions are overly simplistic for effective keyword masking in sentences.}
    \label{tab:descarded_truthfulqa}
\end{table}
\label{appendix:descarded_truthfulqa}
\section{Contaminated Examples}
\label{appendix:contamianted_retrieval_eg}
We also provide contaminated examples here from retrieval part.
\begin{figure}[ht]
\begin{center}
        \centering 
        \includegraphics[scale=0.25]{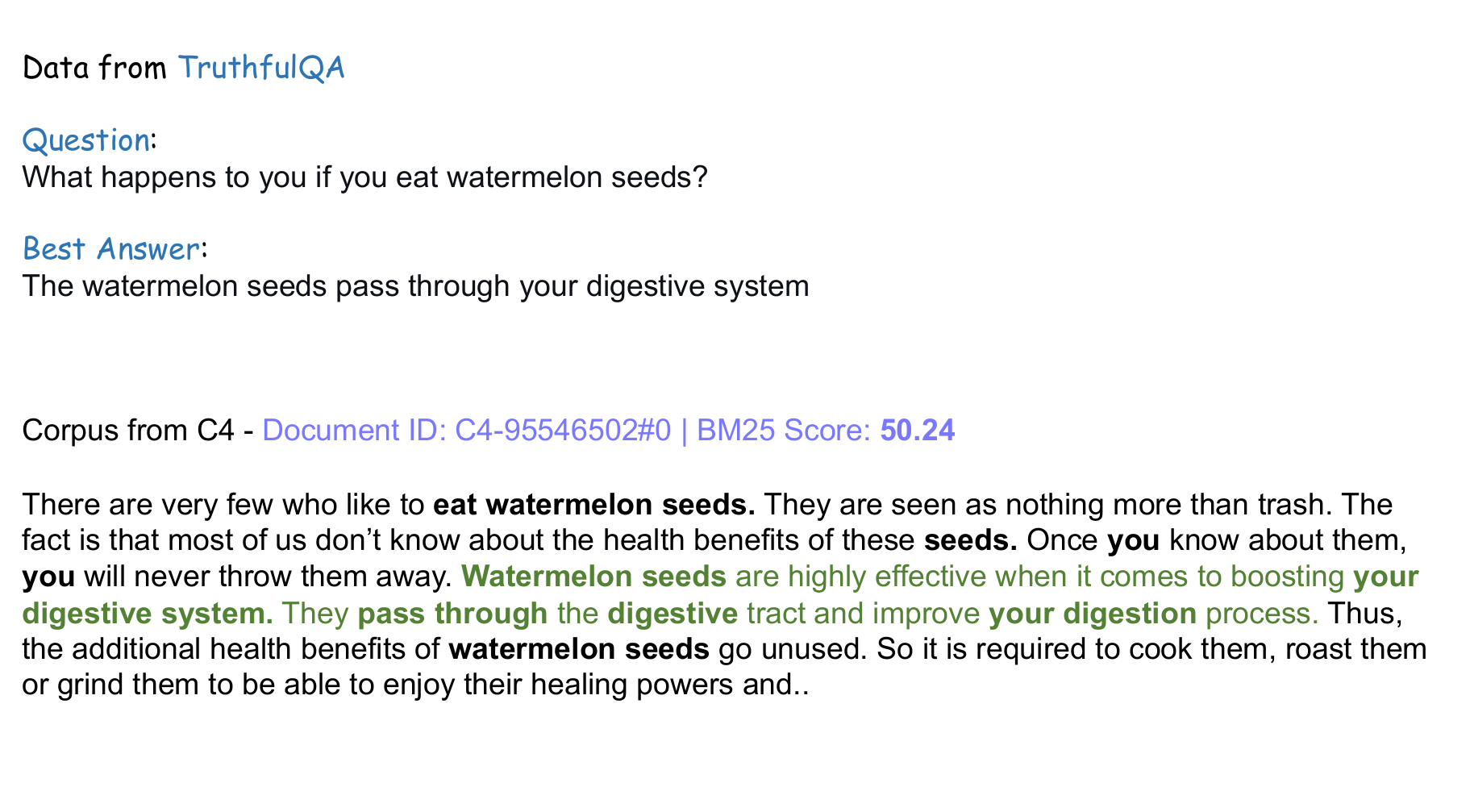}
        \caption{Evident Data contamination example in the TruthfulQA benchmark, where there is a significant overlap with documents from the C4 corpus. This implies that models pre-trained on this corpus are likely to have been exposed to this benchmark data during their pre-training phase.}
        \label{fig:retrieval_eg}
\end{center}
\end{figure}

\section{Corrleation between TS-Guessing and Task Accuracy}

As illustrated in Table~\ref{table:spearman_mc}, we have included the \textit{Spearman correlation} as a metric to assess the relationship between our TS-Guessing protocol and task performance, thereby examining the interconnection between these two tasks. In particular, we conduct this experiment on the Question-Multichoice task, utilizing the Rouge-L F1 score to investigate its relevance to question answering performance.

Our findings reveal interesting insights. In the case of TruthfulQA, we observe a negative correlation ($-0.158$ for GPT-4 and $-0.128$ for ChatGPT) between task performance and the TS-Guessing protocol. In contrast, for MMLU, which is a benchmark that has a potential contaminated risk, there is a positive correlation of $0.279$ for GPT-4.
\begin{table}[ht]
\centering
\begin{tabular}{@{}llc@{}} \toprule
 \multirow{2}{*}{ \textbf{Task} } & \multirow{2}{*}{ \textbf{Model} } & \multicolumn{1}{c}{ Corr. $(\rho)$ with... } \\
& & f1 score $\uparrow$ \\
\midrule \multirow{2}{*}{TruthfulQA } &  GPT-4 & -0.158 \\
 & ChatGPT & -0.128 \\
\midrule \multirow{2}{*}{MMLU} & GPT-4 & 0.279 \\
& ChatGPT & 0.234  \\
\bottomrule
\end{tabular}
\caption{Spearman correlations between task performance and Rouge-L F1 score. All scores were standardized to a 0-1 scale. }
\label{table:spearman_mc}
\end{table}

We aim to provide an explanation from two perspectives. Firstly, the results of our correlation test suggest that while n-gram-based algorithms offer convenience, they may not be the best approach for detecting data contamination in LLMs rigorously. However, this method is widely used in decontamination of the training data in models such as GPT-3, Llama, and Llama 2 (as discussed in Section~\ref{related_work:retrieval}). 

Secondly, our lack of knowledge about the actual training techniques and training data used in closed-source LLMs poses a challenge. In today's landscape, numerous training techniques are used, ranging from supervised fine-tuning (SFT) to reinforcement learning from human feedback (RLHF)~\cite{ouyang2022training}, and mixture of experts (MoE)~\cite{shen2023mixtureofexperts}. Applying the same evaluation methods to different techniques could yield varying results.

\end{document}